\documentclass[%
 superscriptaddress,
 reprint,
 showpacs,preprintnumbers,
 nofootinbib,
 amsmath,amssymb,
 aps,
 prd,
 floatfix,
]{revtex4-2}

\usepackage{graphicx}
\usepackage{nameref}
\usepackage{xcolor}
\usepackage{adjustbox} 
\usepackage{hyperref}
\usepackage{soul}

\usepackage{amssymb,amsmath,amstext,amsthm,amsfonts,amsfonts}
\usepackage[utf8]{inputenc}
\usepackage[ruled]{algorithm2e}
\usepackage{url}
\usepackage{soul}
\usepackage{bm}
\usepackage[normalem]{ulem}
\usepackage{tabularx}
\usepackage{multirow, makecell}
\usepackage{floatrow}
\usepackage{comment}

\usepackage{algpseudocode}

\usepackage{color, colortbl}
\definecolor{Gray}{gray}{0.9}
\definecolor{LightCyan}{rgb}{0.88,1,1}
\usepackage[first=0,last=9]{lcg}

\usepackage{amsmath}

\newfloatcommand{capbtabbox}{table}[][1.275\linewidth]

\newcommand{\affiliationETHZ}{ETH Zurich, Institute for Particle Physics and Astrophysics, CH-8093 Zurich, Switzerland.}
\newcommand{\affiliationSBU}{State University of New York at Stony Brook, Department of Physics and Astronomy, Stony Brook, New York, USA.}

\DeclareMathAlphabet{\mathsfit}{\encodingdefault}{\sfdefault}{m}{sl}
\SetMathAlphabet{\mathsfit}{bold}{\encodingdefault}{\sfdefault}{bx}{n}

\begin{document}

	\title{Deep-learning-based decomposition of overlapping-sparse images:\\ application at the vertex of neutrino interactions} 

    \author{Sa\'ul Alonso-Monsalve}
    \email[E-mail: ]{salonso@ethz.ch}
    \author{Davide Sgalaberna}
    \author{Xingyu Zhao}
    \author{Adrien Molines}
    \affiliation{\affiliationETHZ}
    \author{Clark McGrew}
    \affiliation{\affiliationSBU}
    \author{Andr\'e Rubbia}
    \affiliation{\affiliationETHZ}

\begin{abstract}
\noindent 

\section*{abstract}

Image decomposition plays a crucial role in various computer vision tasks, enabling the analysis and manipulation of visual content at a fundamental level. Overlapping images, which occur when multiple objects or scenes partially occlude each other, pose unique challenges for decomposition algorithms. The task intensifies when working with sparse images, where the scarcity of meaningful information complicates the precise extraction of components. This paper presents a solution that leverages the power of deep learning to accurately extract individual objects within multi-dimensional overlapping-sparse images, with a direct application in high-energy physics with decomposition of overlaid elementary particles obtained from imaging detectors.
In particular, the proposed approach tackles a highly complex yet unsolved problem: identifying and measuring independent particles at the vertex of neutrino interactions, where one expects to observe detector images with multiple indiscernible overlapping charged particles. By decomposing the image of the detector activity at the vertex through deep learning, it is possible to infer the kinematic parameters of the identified low-momentum particles - which otherwise would remain neglected - and enhance the reconstructed energy resolution of the neutrino event.
We also present an additional step - that can be tuned directly on detector data  - combining the above method with a fully-differentiable generative model to improve the image decomposition further and, consequently, the resolution of the measured parameters, achieving unprecedented results. This improvement is crucial for precisely measuring the parameters that govern neutrino flavour oscillations and searching for asymmetries between matter and antimatter.
\end{abstract}

\maketitle

\section{Introduction}
\label{sec:introduction}

Breaking down an image into its constituent components or layers, such as textures, colour channels, shading, or illumination, is known as image decomposition. It is a broad research area within computer vision and image processing with the main goal of extracting meaningful information from an image and separating it into different elements, which can be useful for various applications like denoising, image editing, object recognition, and scene understanding~\cite{Li_2018_CVPR, 6803963, Monnier_2021_ICCV, 9483837}. Techniques such as Fourier Transformations or Principal Component Analysis (PCA), respectively, are traditionally used for decomposing images into their frequencies or orthogonal components~\cite{10.1007/978-3-319-14442-9_20,4383720, 10.1145/2663761.2664195, Strubbe:14, https://doi.org/10.1049/iet-ipr.2018.5499}. The use of deep learning for image decomposition is becoming an active area of research, too, with recent developments focused on developing new techniques and improving the accuracy and efficiency of the process~\cite{8931240, 8354257, Gandelsman_2019_CVPR}.

Furthermore, a ``sparse'' image is a visual representation that exhibits a relatively small percentage of filled or meaningful pixels compared to the total number of pixels, implying a significant fraction of the image is empty or irrelevant. Sparsity can occur for several reasons, such as the natural characteristics of the data, data acquisition processes, or deliberate compression techniques. It is common in several scientific fields, such as cosmology, particle physics, medical imaging, and molecular biology, and putting efforts to deal with such data is crucial for advancing our knowledge~\cite{PhysRevD.103.036012, PMID:33194545,genes14020403, zeng2023medical}. 
Some current image-decomposition solutions have been designed for sparse images; however, these methods tend to target non-scientific data since their final goal is to solve tasks such as image editing, compression or denoising~\cite{mou2022compressed, ManjuLeninFred+2020+515+528, LI2021116281, DU2023109241, liu2021hyperspectral}. The decomposition of overlapping-sparse images is an extremely complex task due to the inherent ambiguities, particularly in scenarios like overlaying tracks in a particle-physics event. In such cases, the challenge lies in disentangling the individual signal contributions and accurately reconstructing the underlying composition. The problem becomes even more intricate in cases where the number and type of tracks can vary, resulting in a vast number of potential candidates. Attempting to explore all possible combinations exhaustively becomes infeasible due to the rapidly expanding search space, rendering traditional computational methods impractical.

Deep learning offers a promising solution to this problem. By training neural networks on large datasets of known image configurations, deep-learning models can learn complex patterns and relationships within overlapping images. They can capture the underlying physical and statistical properties, enabling them to infer the constituent elements from the overlapping components. The advantage of deep learning lies in its ability to efficiently handle high-dimensional data and extract relevant features for accurate decomposition automatically. There has been some recent exploration in this area, particularly in utilising deep learning to decompose geophysical images or to recover events that are distorted by multiple collisions, known as pile-ups~\cite{2021RG000742, komiske2017pileup, KIM2023168492}.
However, these approaches are limited in their applicability as they make prior assumptions to simplify the problem. These assumptions include a fixed number of overlapping elements and images of fixed size, which restrict their generic nature.

The hypothesis we propose suggests that the transformer model~\cite{vaswani2017attention, 10278387}, an architecture that is undeniably reshaping the landscape of deep learning and powering exceptional chatbots like ChatGPT~\cite{chatgpt}, possesses the ability to grasp pixel correlations among varied images that intersect within a limited 3D space, utilising its attention mechanisms through a self-supervised scenario. Furthermore, the decoding component inherent to the transformer model holds the potential to progressively deconstruct images by internally subtracting the predicted individual images in an iterative manner.

\begin{figure*}[htb] 
\centering
  \includegraphics[width=0.85\textwidth]{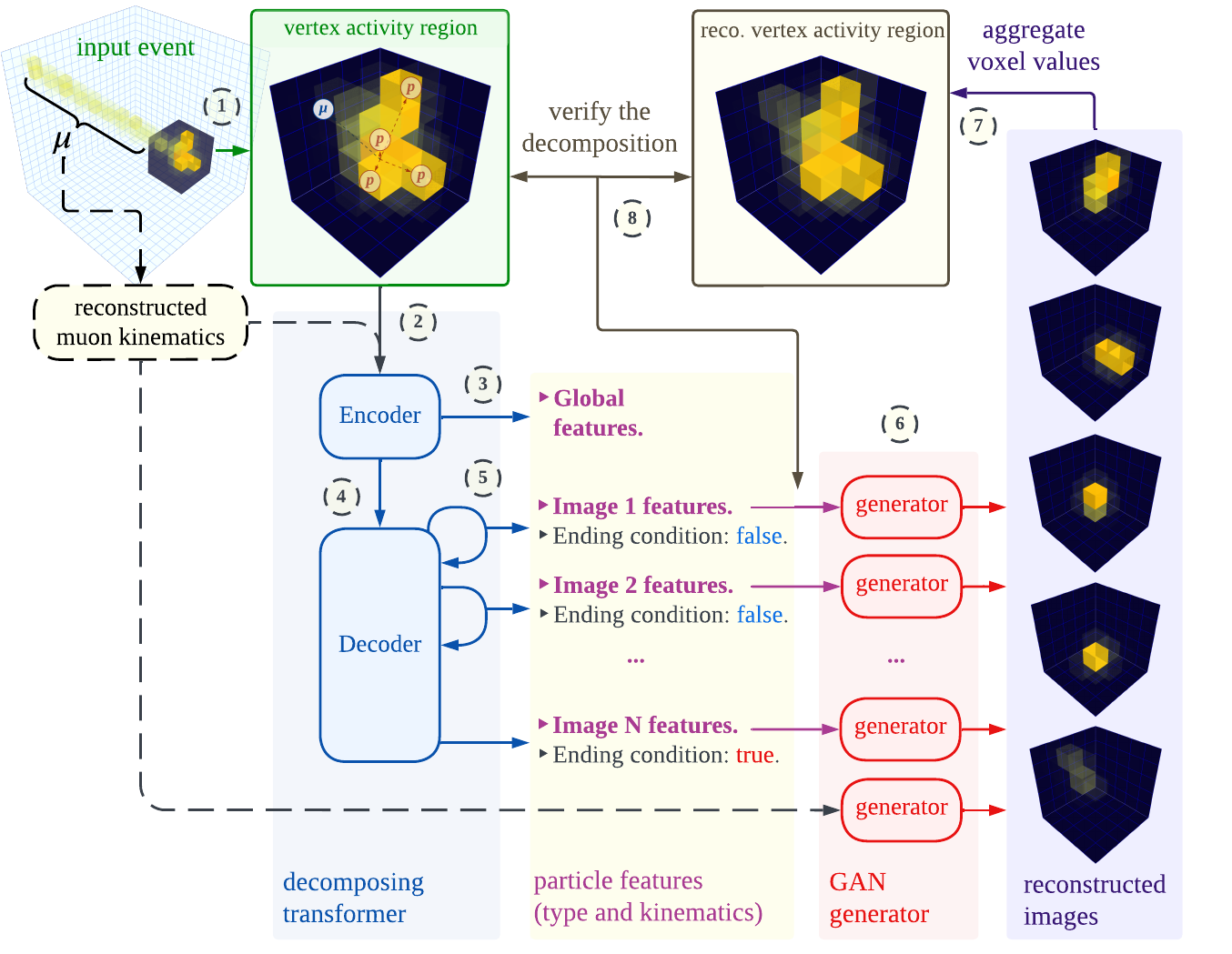}
  \caption{\label{fig:workflow}
  \textbf{Pipeline for an example neutrino interaction.} (1) Detection of the vertex activity (VA) region in the input event. The zoomed-in view unveils the particle mosaic at the interaction vertex. (2) Utilisation of the VA image, combined with the reconstructed kinematic parameters of the escaping muon, as input for the transformer encoder. The transformer encoder processes the input, resulting in (3) the reconstruction of the interaction vertex position and (4) an embedded representation of the VA event for the decoder. (5) The transformer decoder initially processes the encoder's information, resulting in a prediction for the kinematics of the most energetic particle in the input VA. Simultaneously, it generates a boolean variable to signify the existence of additional particles that require reconstruction. In cases where additional particles are identified, the transformer decoder proceeds to iteratively provide their kinematic predictions in descending order of the kinetic energies of the particles. This process continues by incorporating encoder data and the kinematics of the previously predicted particle until the boolean variable signals termination, indicating that the transformer has determined no further particles are present in the VA event. (6) A generative adversarial network (GAN) generator produces images of particles based on the kinematics predicted by the transformer. Using the initial reconstructed kinematics, the GAN is also employed to generate an image of the escaping muon. (7) The generated images are aggregated by summing their voxel photoelectrons, and (8) compared with the input VA event to verify the decomposition process. The workflow may return to step 6 to further optimise kinematic parameters.}
\end{figure*}

This paper presents a generic solution that leverages deep-learning techniques for the decomposition of overlapping-sparse images in the scientific domain. The proposed method is applied to a challenge in particle physics: resolving degeneracies at the vertex of neutrino interactions to comprehend the properties of the particles involved. When a neutrino interacts within a scintillating target, it generates secondary charged particles that produce scintillation light. Specifically, we take the case: 

\vspace{-0.1cm}
\begin{equation}
\nu_{\mu}+N\rightarrow\mu^{-}+X
\end{equation}

where $N$ is the interacting nucleon and $X$ is the so called hadronic system. Within the finite granularity of the detector, some of these particles exhibit indistinct trajectories, leading to the production of coincidental scintillation signals, which can introduce ambiguity in the data.

Resolving such signals with percent-level energy resolution
will play a critical role in the future high-precision 
long-baseline accelerator neutrino-oscillation experiments, such as DUNE \cite{Abi:2020qib} and Hyper-Kamiokande \cite{Abe:2018uyc}, directly impacting the sensitivity to the potential discovery of leptonic charge-parity ($CP$) violation and the measurement of the neutrino oscillation parameters, including the $CP$ violating phase, the neutrino mass squared difference ($|m^2_3 - m^2_2|$), and the mixing angle $\theta_{23}$~\cite{t2k2020constraint}.
Accurate neutrino energy reconstruction is pivotal for precise measurement of oscillation probabilities, as any mismodelling can introduce biases in inferred oscillation parameters.
One source of systematic uncertainty is given by the so-called ``vertex activity'', which refers to the energy loss by low-momentum final-state charged hadrons that leave overlapping indiscernible signatures in the detector near the neutrino interaction point. This phenomenon results in energy loss that is challenging to attribute to individual particles and, consequently, to the original neutrino energy
\cite{ershova2023role, PhysRevD.106.032009}. 
The key role is played by the quenching process of the generated scintillation light
\cite{birk1,BIRKS196439}, 
exhibiting variations based on the type and energy of the ionising particle.
Despite efforts in current accelerator neutrino experiments to measure vertex activity
\cite{MINERvA:2015ydy,MINERvA:2018nab,MINERvA:2021wjs,MINERvA:2022mnw}, the absence of suitable reconstruction and analysis tools makes measurements subject to model dependence, wherein an arbitrary selection of the nuclear model for the unfolding process introduces biases in the overall neutrino energy reconstruction. Therefore, developing solutions that facilitate the precise unfolding of scintillation quenching by accurately estimating the number, type, and energy of final-state particles becomes crucial.

In this study, we present a novel methodology to decompose multi-dimensional images into distinct and independent objects. To the best of our knowledge, our tool represents the first of its kind, enabling the inference of crucial parameters such as the number, type, and energy of final state hadrons within the vertex activity region of neutrino interactions, all without reliance on arbitrary a-priori nuclear models. Leveraging deep neural networks, our approach facilitates this decomposition efficiently, overcoming computational barriers that were previously prohibitive. This work employs a realistic particle detector simulation to demonstrate how to use a transformer to reconstruct the kinematics of each produced particle and extract independent particle images.
Moreover, the article showcases the validation and improvement of the proposed process on detector data, employing a generative model and a comprehensive parameter space exploration through gradient-descent minimisation, exploiting the full differentiability of the model. The selection of a combination of particles that minimises a likelihood function further solidifies the effectiveness and reliability of the presented approach.

\section{Results}
\label{sec:results}

Figure~\ref{fig:workflow} depicts the image-decomposition workflow designed for and applied to images of the region around the vertex of neutrino interactions, known as vertex activity, where independent particle tracks are not easily discernible. The input images consist of overlapping simulated particles, mainly hadrons, releasing their energy in the proximity of the neutrino interaction vertex in a voxelised 3D detector (details in Section~\ref{sec:simulation}). The non-empty voxels represent the energy loss (in number of photoelectrons) by the particles in the detector, meaning that multiple particles can contribute to the same voxel, making it challenging to determine the number of photoelectrons that correspond to each particle or guess the number of particles that contribute to a particular interaction vertex. A transformer neural network is run on the overlapping image (see Section~\ref{sec:transformer}, for more details). It is used to extract both global features of the image (in the neutrino-interaction case, an accurate position of the interaction vertex) and some characteristics of each of the independent images involved: in our physics scenario, the transformer outputs iteratively for each reconstructed particle its type, some kinematics (i.e., initial kinetic energy and track direction), and a boolean value indicating whether to stop. This iterative process follows a scheme where the particles are reconstructed based on their kinetic energy in descending order; it stops when there are no particles left. In that way, the most-energetic particles, which are the ones that contribute most to the total energy loss in the event, are reconstructed first. In contrast, low-energetic particles are left for the end. A generator is used at this stage to create images of independent particles using the information reconstructed by the transformer as input. The choice of the generator can range from a classic simulation to a fully-differentiable one (e.g., in the form of a generative model or a differentiable simulation such as \cite{gasiorowski2023differentiable, 9812293, dorigo2022endtoend}). In this article, the generator developed is a generative adversarial network (GAN)~\cite{10.5555/2969033.2969125} (more information in Section~\ref{sec:gan}), a widely acknowledged algorithm known for its rapid image generation capabilities. It was used to verify the transformer decomposition and perform an additional optimisation of the reconstructed kinematics to further improve the decomposition. The process operates as follows: for a specific event, a minimisation algorithm is applied to further refine the parameters generated by the transformer. The individual voxel photoelectrons from the GAN-generated particle images are summed, creating a composite reconstructed image that is subsequently compared to the original image to minimise a chosen target metric, such as a loss or a likelihood function.

\subsection{Initial case}
\label{sec:initial_case}

The conducted analysis focused on a scenario involving exclusively multi-particle final states analogous to $\nu_{\mu}$ charged-current (CC) interactions with no outgoing pions in the detector (CC$0\pi$ topology), the most typical interaction mode of neutrino experiments like T2K~\cite{ABE2011106} or MicroBooNE~\cite{Acciarri-2017-design}, both having the neutrino spectrum mostly below 1~GeV.
Each event in our investigation was characterised by an energetic muon and a varying number of protons, ranging from 1 to 5, produced at the interaction vertex and dispersing along distinct directions with different energies. A comprehensive breakdown of the configurations employed in our study is presented in Section~\ref{sec:simulation}.

\begin{figure*}[htb] 
\centering
  \adjincludegraphics[width=1.0\textwidth, trim={0cm 0 0 0},clip]{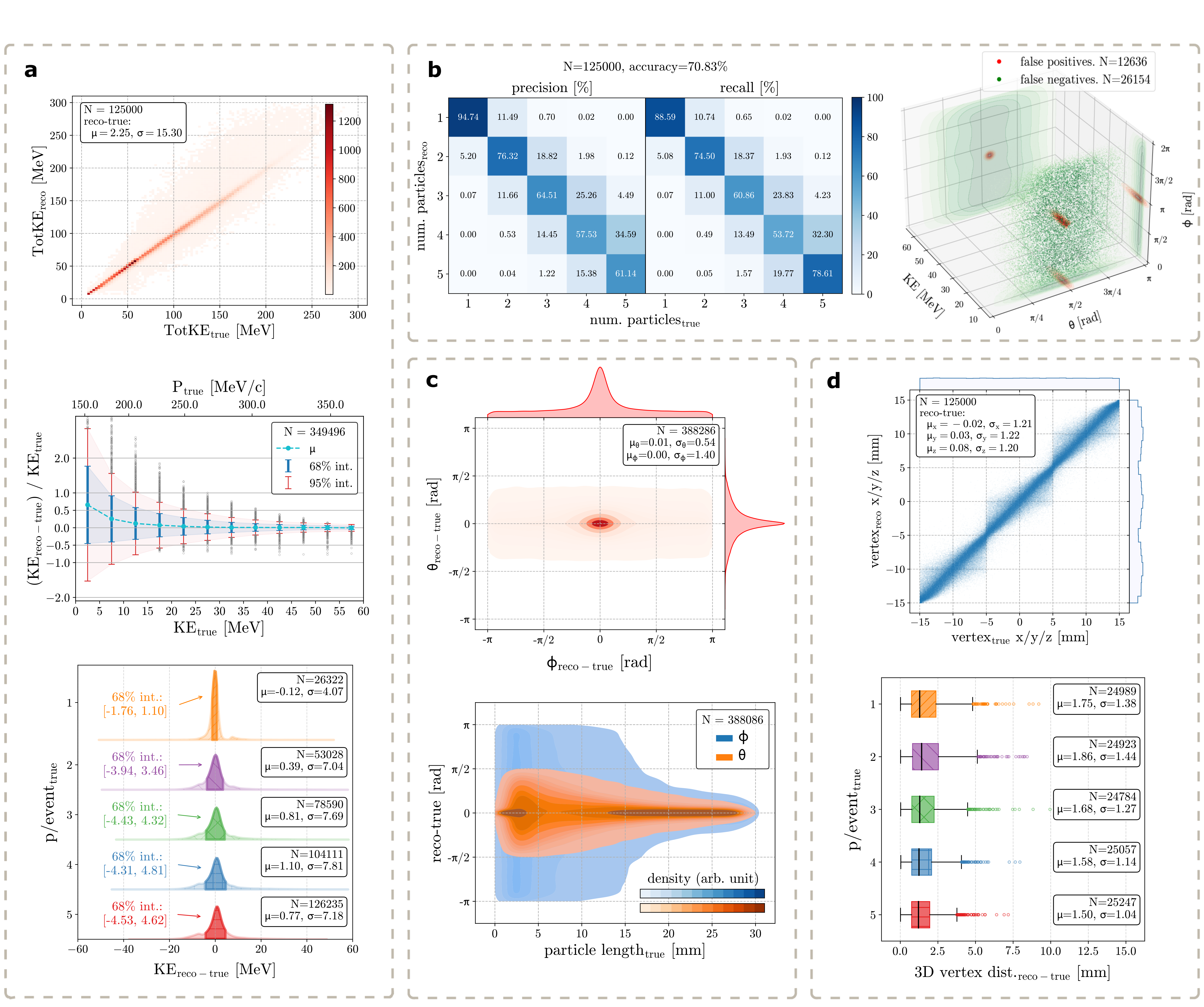}
  \caption{\label{fig:results}
  \textbf{Main results of the vertex-activity (VA) fitting algorithm on the testing dataset, consisting of events with one muon and one to five protons.} $\mu$: mean, $\sigma$: standard deviation, false positives (negatives): over (under)-reconstructed particles (with default true (reconstructed) values: kinetic energy (KE) = 0 MeV; $\theta=$ 90 degrees; $\phi=$ 180 degrees).
  \textbf{a} Top: histogram of the total vs reconstructed KE for each event. The plot includes false positives and false negatives. Middle: KE resolution per particle relative to true KE and momentum (P) for each fitted particle. The plot excludes false positives and false negatives. Bottom: distribution of the difference between the reconstructed KE and the true KE per particle for the cases with 1-5 particles per event. The plot includes false positives and false negatives. 
  \textbf{b} Left: confusion matrix showing the recall (normalisation by columns) and precision (normalisation by rows) of the true vs reconstructed number of particles. Right: distribution of the false positives (particles predicted by the algorithm that are not present in the input events) and false negatives (particles not predicted by the algorithm that are present in the input events).   
  \textbf{c} Top: density plot relating the difference between reconstructed and true $\phi$ and $\theta$ (spherical coordinates) per particle. Bottom: density plot relating the difference between reconstructed and true angles and the particle length. Both plots include false positives and false negatives. 
  \textbf{d}. Top: scatter plot comparing the true and reconstructed vertex per event for the x, y, and z coordinates. Bottom: box plot of the 3D Euclidean distance between the reconstructed and true vertices for the cases with 1-5 particles per event. }
\end{figure*}

\begin{figure*}[htb] 
\centering
  \adjincludegraphics[width=0.90\textwidth, trim={0cm 0 0 0},clip]{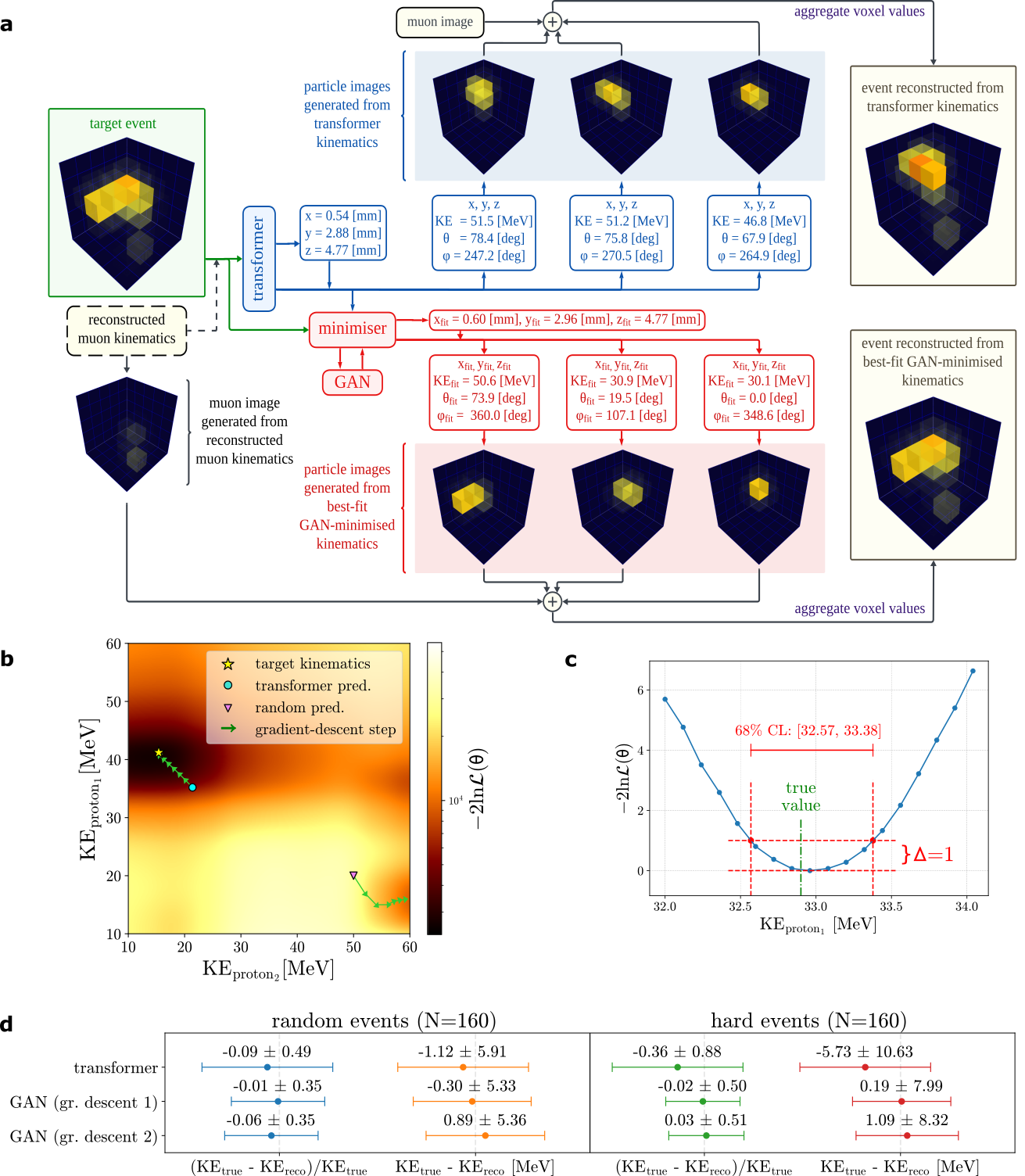}
  \caption{\label{fig:minimiser}
  \textbf{Kinematic parameter optimisation via gradient-descent minimisation.}
  \textbf{a} Processing of a target event: The target event, along with its reconstructed muon kinematics, is input to the transformer (depicted in blue). The transformer generates a set of possible kinematic combinations for all particles within the target event. These kinematics are subsequently forwarded to the gradient-descent minimiser (depicted in red), which leverages the generative adversarial network (GAN) to refine the kinematics and improve the correspondence with the target event. The diagram visually represents the different decompositions resulting from this process. Kinematic parameters: kinetic energy (KE), direction in spherical coordinates ($\theta$ and $\phi$), interaction vertex 3D position (x, y, z). \textbf{b} The plot shows two scenarios in a likelihood space (pre-computed for the kinetic energy of two most energetic protons of an arbitrary event: KE$_{\text{proton}_{1}}$ and KE$_{\text{proton}_{2}}$). One starts from the transformer output and successfully reaches the target values, while the other begins at a random parameter space point and gets stuck in a local minimum. \textbf{c} Profiled negative log-likelihood $\mathcal{L}$ for the kinetic energy of the most energetic proton (KE$_{\text{proton}_{1}}$) of an arbitrary event, and the curve shows the 68\% confidence interval determined by a $\Delta\mathcal{L}$ of 1 for one degree of freedom. \textbf{d} The resolution of kinetic energy (KE), as determined through an analysis of sets of random and ``hard'' events (i.e., events where the image reconstructed from the transformer exceeded a predefined mean-squared-error threshold in comparison to the target image), was assessed for three distinct methodologies: the transformer and two gradient-descent techniques (``GAN (gr. descent 1)'' and ``GAN (gr. descent)'', as per Algorithms~\ref{alg:gradient_descent1} and~\ref{alg:gradient_descent2} from Section~\ref{sec:gradient_descent}, respectively). It illustrates the effectiveness of GAN-based minimisation in refining the kinematic parameters.}
\end{figure*}

The performance of the decomposing transformer is presented in Fig.~\ref{fig:results}. One of the most striking outcomes of our study is the precision with which kinetic energy (KE) is reconstructed, both on a per-particle basis and in terms of the total KE per event (Fig.~\ref{fig:results}a). This is particularly evident for high-energy protons, which constitute the predominant contributors to the cumulative charge of the input vertex-activity images. This precision in reconstructing KE plays a pivotal role in accurately characterising the underlying physics in neutrino interactions. 
Besides, the estimation of the total number of protons within the vertex-activity region demonstrates an accuracy exceeding 70\%.
However, when considering Fig.~\ref{fig:results}b, it is evident that most under- and over-reconstructed protons occur in the lowest KE range. This feature can be attributed to protons with low KE contributing negligibly to the total energy loss in the event, and those protons tend to traverse distances not beyond the confines of a single voxel. Notably, by adopting a criterion allowing for an error of up to ±1 reconstructed protons, the accuracy of our estimation surpasses 98\%, underscoring the effectiveness of our methodology. 
The reconstruction of the direction in spherical coordinates exhibits a strong dependency on the length of the particle track (Fig.~\ref{fig:results}c), which is influenced by its KE. As a consequence, the precision of the reconstructed direction significantly improves with longer particles. Conversely, particles failing to escape the initial voxel present challenges in terms of direction reconstruction due to their confinement within the detection volume. The spatial resolution achieved in vertex reconstruction is a remarkable benchmark (Fig.~\ref{fig:results}d), amounting to approximately 2 mm. This resolution is superior to the intrinsic resolution of the examined detector, comprised of voxels with dimensions of 1 cm$^3$. 
The exceptional vertex position resolution underscores the capabilities of our approach in pinpointing the origin of neutrino interactions with remarkable precision.

\subsection{Leveraging the generator for an enhanced decomposition}
\label{sec:gan}

\begin{figure*}[htb] 
\centering
  \adjincludegraphics[width=1.0\textwidth, trim={0cm 0 0 0},clip]{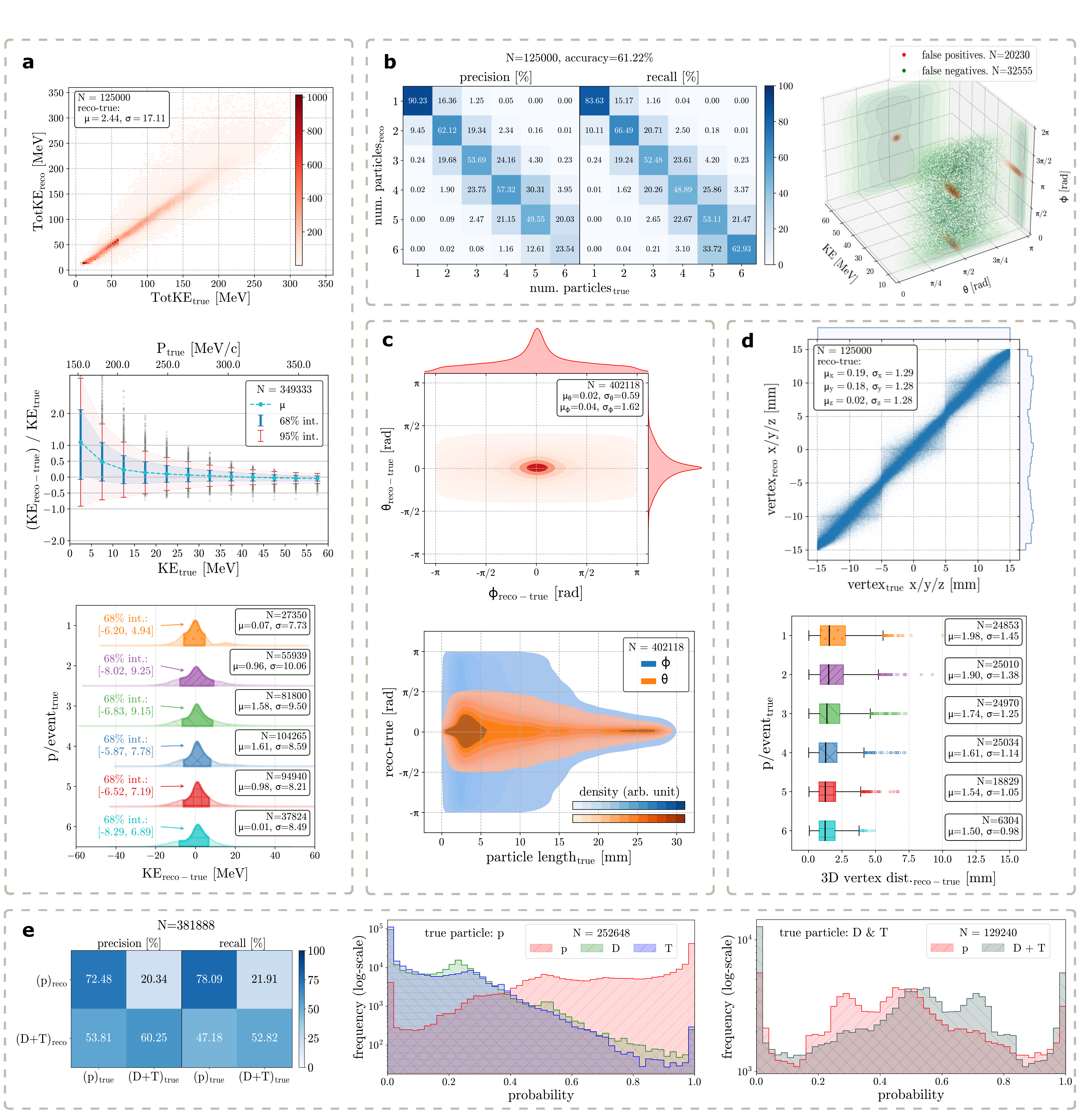}
  \caption{\label{fig:results_clusters}
  \textbf{Main results of the vertex-activity (VA) fitting algorithm on the testing dataset, consisting of events with one muon, 0-4 protons (p), 0-1 deuterium (D), and 0-1 tritium (T).} $\mu$: mean, $\sigma$: standard deviation, false positives (negatives): over (under)-reconstructed particles (with default true (reconstructed) values: kinetic energy (KE) = 0 MeV; $\theta=$ 90 degrees; $\phi=$ 180 degrees).
  \textbf{a} Top: histogram of the total vs reconstructed KE for each event. The plot includes false positives and false negatives. Middle: KE resolution per particle relative to true KE and momentum (P). The plot excludes false positives and false negatives. Bottom: distribution of the difference between the reconstructed KE and the true KE per particle for the cases with 1-6 particles per event. The plot includes false positives and false negatives. 
  \textbf{b} Left: confusion matrix showing the recall (normalisation by columns) and precision (normalisation by rows) of the true vs reconstructed number of particles. Right: distribution of the false positives (particles predicted by the algorithm that are not present in the input events) and false negatives (particles not predicted by the algorithm that are present in the input events).   
  \textbf{c} Top: density plot relating the difference between reconstructed and true $\phi$ and $\theta$ (spherical coordinates) per particle. Bottom: density plot relating the difference between reconstructed and true angles and the particle length. Both plots include false positives and false negatives. 
  \textbf{d}. Top: scatter plot comparing the true and reconstructed vertex per event for the x, y, and z coordinates. Bottom: box plot of the 3D Euclidean distance between the reconstructed and true vertices for the cases with 1-6 particles per event.
  \textbf{e} Left: confusion matrix showing the recall (normalisation by columns) and precision (normalisation by rows) of the particle identification. Middle: particle-identification probability distributions when the true particle is a proton. Right: particle-identification probability distributions when the true particle is a nuclear cluster (either deuterium or tritium).}
\end{figure*}

The output of our decomposing transformer presents a groundbreaking solution to reconstruct the vertices of neutrino interactions. As described previously, these findings serve as invaluable input for a generator capable of producing images for each reconstructed particle, which allows us to verify the decomposition process by comparing the target input image with the aggregated version of the reconstructed particles, as illustrated in steps 6, 7, and 8 of Fig.~\ref{fig:workflow}.
We can further enhance our reconstruction by deploying a minimisation algorithm seeded by the transformer output. This algorithm leverages the GAN generator to create images for each particle, facilitating the search for combinations of particles (and, thus, their kinematics) that better match the target event. In our implementation, the generator is a fully differentiable generative model - a generative adversarial network - efficiently enabling us to perform gradient-descent optimisation on the kinematic properties using standard deep-learning frameworks, otherwise unfeasible using conventional simulations. The approach is shown in Fig.~\ref{fig:minimiser}a.

In our study, we have devised two distinct variants of the gradient-descent optimisation approach applied to particle kinematics for a specific target event, where the output of the transformer is utilised as the input initial kinematic parameters to optimise. These variants are contingent upon treating the stochastic component of the generative model, with the model having learnt to correlate noise seeds with the random fluctuations observed in the training dataset. The initial variant aims to identify the optimal set of kinematic parameters that closely align with the target image while simultaneously determining the most suitable random input for the GAN generator. This selection process involves running $n$ independent gradient-descent runs, each minimising a loss, typically represented as the mean-squared error between the generated and target images. Each run employs a distinct input noise seed, and the run that results in the lowest loss is chosen to determine the kinematic parameters, thereby achieving the best fit with the target image. In contrast, the second variant seeks to strike a balance by disregarding the best random input and prioritising the ability to provide precise error metrics for subsequent physical analyses. In this variant, a single gradient-descent run is conducted, and $n$ GAN-generated images are averaged and compared to the target image in each iteration to minimise a log-likelihood. The technical details of both variants are described in Section~\ref{sec:gradient_descent}.

As depicted in Fig.~\ref{fig:minimiser}b, we observe the inherent utility of utilising the transformer output as an initial point for the gradient-descent process. This choice prevents the algorithm from becoming stuck in local minima, thus obviating the necessity for a more intricate learning-rate scheduling strategy that may not necessarily yield preferable outcomes. In addition, the second variant allows for the profiling of the chosen likelihood and the computation of confidence intervals for individual kinematic parameters while treating the other parameters as nuisances, as visualised in Fig.~\ref{fig:minimiser}c. This step is pivotal in the process of quantifying uncertainties, which, in turn, facilitates the effective incorporation of our findings into extended data analyses. Besides, the additional refinement of the kinematic parameters intelligibly improves their correctness, as demonstrated in Fig.~\ref{fig:minimiser}d, where both gradient-descent variants consistently outperform the results achieved by the transformer model, particularly when dealing with cases in which the image reconstructed from the transformer output deviates significantly from the desired target.

\subsection{Nuclear clusters}
\label{sec:nuclear_clusters}

The most complex scenario we investigated in our study pertains to the presence of nuclear clusters, as discussed in~\cite{PhysRevD.106.032009, ershova2023role}. These clusters, composed explicitly of deuterium and tritium nuclei in our studies, in addition to the concurrent existence of muons and protons within our vertex activity, present a formidable challenge. Their composite nature introduces inherent complexities into the process of decomposition for several distinct reasons. One primary challenge lies in the intrinsic ambiguities that they introduce, which makes distinguishing these composite nuclear clusters from combinations of independent protons a particularly intricate endeavour. In fact, in certain instances, it becomes practically unfeasible to achieve such differentiation.
In this scenario, the decomposing transformer model is confronted with the task of not only accurately inferring their kinematic properties but also deciphering the specific particle types. The interplay of these various components adds layers of complexity to the decompositional process, making it a crucial and challenging area of investigation within our research.

We revisited our original analysis, wherein each event comprised a muon, a variable number of protons ranging from 0 to 4, and added the possibility of including 0 to 1 deuterium and 0 to 1 tritium nuclei. We maintain the criterion that each event must include at least one particle in addition to the muon and not exceed a maximum total of seven particles.

In Fig.~\ref{fig:results_clusters}, we reproduce the same plots as presented previously, but now focusing on the case involving nuclear clusters. The reconstruction of kinetic energy (Fig.~\ref{fig:results_clusters}a) appears to be minimally affected, with any slight degradation in performance attributable to the increased complexity of events in this scenario. Notably, we observe a discernible reduction of nearly 10\% in the accuracy of particle number estimation (Fig.~\ref{fig:results_clusters}b), with most under- and over-reconstructed protons concentrated within the lowest kinetic energy range as before. The decline in performance can be readily attributed to the increased number of particles in these events, aggravated by the inherent complexity of the nuclear clusters. Nevertheless, if we adopt a criterion allowing for an error of up to ±1 reconstructed particles as before, our estimation accuracy exceeds 96\%. Similar observations can be made regarding the reconstruction of the vertex position (Fig.~\ref{fig:results_clusters}d). The inclusion of additional particles, rather than facilitating the reconstruction, results in the emergence of a small energy cluster around the interaction vertex, moderately complicating the precise determination of its location. Additionally, the reconstruction of the direction of the added nuclear clusters proves to be a challenging task, as evidenced by the accumulation of incorrectly reconstructed directions within lower energy ranges (Fig.~\ref{fig:results_clusters}c). The particle identification output exhibits remarkable performance, as evidenced in Fig.~\ref{fig:results_clusters}e. The network displays a notable level of accuracy, with a correct identification rate exceeding 78\% for protons. Identification errors are infrequent when the output probability exceeds 0.8, stressing the robustness of the method. On the contrary, difficulties arise when differentiating nuclear clusters from short protons despite successfully identifying over half of the nuclear clusters. As expected, distinguishing nuclear clusters from short protons leads to significant ambiguities in the identification process.

In summary, the decompositional performance remains impressive. This approach can be regarded not only as a proof-of-concept but also as a potent method, demonstrating its potential for successful analyses when appropriately trained with the correct assumptions.

\subsection{Comparison with standard method}
\label{sec:comparison}

The vertex-activity (VA) measurements reported in literature predominantly suffer from ``model dependence'', meaning that an arbitrary choice of a specific neutrino interaction model is used to unfold the truth energy loss of VA hadrons, which typically simplifies to the following formula for the visible energy (VisE) of the VA, considering a single proton responsible for all VisE:

\begin{equation}
\label{eq:std}
VisE_{\text{reco}} [\text{MeV}] = \frac{1}{1 - C_B \frac{E_{cali}}{\Delta X}} \cdot E_{cali}
\end{equation}

where:

\begin{itemize}
\item $C_B$: Birks coefficient, equal to 0.0126 cm/MeV.
\item $E_{cali}$: represents a relation equal to $\frac{E_{\text{loss}}[p.e.]}{c_{\text{cali}}}$, where $E_{\text{loss}[p.e.]}$ is the total deposited energy (in photoelectrons), and $c_{\text{cali}}$ is a calibration factor equal to 100 p.e./MeV.
\item $\Delta X$: approximate length of the longest proton, in millimetres.
\end{itemize}

\setlength{\tabcolsep}{3pt}
\begin{table*}[htb]
    \centering
    \resizebox{1.0\linewidth}{!}{
    \begin{tabular}{crrrrrrrrrrr}
    \hline \hline
    \multirow{1}{*}{\textbf{true}} & \multirow{2}{*}{\textbf{events}} & \multicolumn{1}{c}{\multirow{1}{*}{\textbf{reco}}} & \multicolumn{3}{c}{\multirow{2}{*}{\textbf{VisE$_{\text{true-reco}}$ (RMSE) [MeV]}}} & \multicolumn{6}{c}{\multirow{1}{*}{\textbf{VisE resolution (RMS) [\%]}}} \\ 
    \textbf{protons} & & \textbf{protons [\%]} & & & & \multicolumn{3}{c}{\textbf{VA region}} & \multicolumn{3}{c}{\textbf{entire event}}\\
    %
    & & transformer & std. method & trans. & trans.+GAN & std. method & trans. & trans.+GAN & std. method & trans. & trans.+GAN\\   
    1 & 36505 & 100.00 & 4.71$\pm$0.02 & 3.85$\pm$0.02 & 2.89$\pm$0.02 & 33.06$\pm$0.25 & 27.41$\pm$0.14 & 26.36$\pm$0.19 & 2.77$\pm$0.01 & 2.38$\pm$0.02 & 1.34$\pm$0.01 \\
    2  & 3520 & 89.35 & 13.35$\pm$0.13 & 6.88$\pm$0.11 & 5.25$\pm$0.09 & 22.44$\pm$0.32 & 16.60$\pm$0.28 & 13.05$\pm$0.22 & 3.79$\pm$0.05 & 2.19$\pm$0.04 & 1.66$\pm$0.03 \\
    3  &  370  & 65.13 & 20.32$\pm$0.58 & 9.29$\pm$0.48 & 7.14$\pm$0.37 & 19.18$\pm$0.56 & 11.06$\pm$0.57 & 8.41$\pm$0.43 & 4.98$\pm$0.17 & 2.42$\pm$0.13 & 1.83$\pm$0.10 \\
    4  &  49  & 65.30 & 27.50$\pm$1.92 & 11.67$\pm$1.38 & 9.51$\pm$1.15 & 20.61$\pm$1.32 & 11.35$\pm$1.35 & 9.34$\pm$1.14 & 6.06$\pm$0.50 & 2.91$\pm$0.35 & 2.32$\pm$0.29 \\

    \hline \hline    
   \end{tabular}
   }
    \caption{
    \label{tab:results} 
    Comparison of the performance between the standard method (Equation~\ref{eq:std}), the transformer model (Section~\ref{sec:initial_case}), and the transformer model together with the generative adversarial network (GAN, Section~\ref{sec:gan}) in reconstructing the visible energy (VisE) across varying proton multiplicities (1-4), 
    within both the vertex-activity (VA) region and the entirety of the event. The resolution is calculated as $(VisE_{true}-VisE_{reco})/VisE_{true}$.}
\end{table*}

While widely adopted, this conventional method introduces systematic uncertainties due to its arbitrary assumptions~\cite{MINERvA:2015ydy,MINERvA:2018nab,MINERvA:2021wjs,MINERvA:2022mnw}. These uncertainties arise from the fact that scintillation light produced by ionising charged particles is affected by Birks' quenching~\cite{birk1,BIRKS196439}, which makes the reconstructed VisE vary as a function of the hypothesised number, type and kinetic energy of final-state particles.

To ensure the reliability of our approach, we tested it on neutrino interactions generated by the NEUT generator~\cite{Hayato_2021}. These events are distinct statistically from the dataset used to train our method, which was crafted to capture the full spectrum of neutrino interaction configurations.
Results are presented in Table~\ref{tab:results}, which compares the reconstructed VisE for the standard method (Equation~\ref{eq:std}) and the proposed alternatives (transformer and transformer+GAN); for the latter, the VisE is calculated by summing the predicted particle kinetic energies. The table demonstrates a pronounced enhancement by the proposed models in every scenario, particularly evident with increasing proton multiplicities. In particular, we get absolute improvements in VisE resolution up to $\sim$12\% in the VA region, and $\sim$4\% for entire events.

\section{Discussion}
\label{sec:discussion}

The results presented in this paper mark a significant advancement in the field of computer vision and its applications in the domain of particle physics. The successful implementation of deep-learning techniques for the decomposition of overlapping-sparse images not only showcases the potential of artificial intelligence in resolving intricate visual content but also opens up new exciting possibilities for addressing complex challenges. By introducing the decomposing transformer, an architecture initially hailing from natural language processing and underpinned by a self-supervised training scheme, this work represents a breakthrough in the analysis and manipulation of multi-dimensional overlapping-sparse images, a problem that has been traditionally problematic for conventional image processing methods.

Employing transformers for sequential tasks is a well-established practice. However, exploiting the iterative nature of transformers to decompose sparse images of overlapping elements introduces a more intricate spatial relationship challenge. We hypothesise that within this iterative process, the transformer internally subtracts the predicted individual components of the image. By reconstructing images based on their most representative features in a certain order (based on the kinetic energy in the neutrino-interaction case), the system prioritises the elements of the image that contribute most significantly to the overall structure. This approach represents a fundamental shift in image decomposition strategies, which often treat all image components equally, regardless of their significance~\cite{9491732,ipol.2011.blmv_ct}. This unique ordering strategy presents several distinct advantages. It closely mirrors the physical reality of various scenarios where specific components dominate the observed event. By prioritising the reconstruction of these influential components, the system efficiently captures elements of the highest interest, a characteristic that holds true in particle physics and many other computer vision applications. Focusing on pivotal components enables the swift creation of a coherent and informative representation of the scene, significantly elevating the efficiency and accuracy of the overall decomposition process.

Introducing a generative model such as a generative adversarial network (GAN) as the image generator represents a significant advancement in integrating computer vision and deep-learning techniques into this problem domain. GANs have demonstrated remarkable proficiency in creating highly realistic and detailed images, making them an ideal choice for generating independent image components based on the information extracted by the transformer. This innovation not only validates the effectiveness of the transformer output but also unveils exciting opportunities for further optimising the reconstructed kinematics. The synergy between the transformer and the GAN, facilitated by the gradient-descent minimisation algorithm, leverages the full differentiability of the GAN. The resultant composite reconstructed image undergoes a rigorous comparison with the original image, primarily focusing on minimising target metrics, such as loss or likelihood functions. This feedback loop, where the system iteratively enhances its decomposition, establishes a mechanism for continuous refinement, enabling greater precision in reconstructing complex and overlapping images. It offers an innovative perspective on the challenges posed by overlapping images, and it can potentially find applications in various fields where image decomposition and reconstruction play pivotal roles, such as medical imaging or environmental monitoring. Additionally, this method has the potential to be applied in mitigating the pile-up of overlapping tracks in particle or nuclear physics, providing a broader spectrum of applications.

One of the most compelling aspects of this research is its direct applicability in the study of neutrino interactions in long-baseline flavour-oscillation experiments. Neutrinos play a fundamental role in our understanding of particle physics and the cosmos at large~\cite{Mohapatra_2007, stecker2023neutrino}. The proposed approach offers a novel perspective on comprehending and extracting valuable information on the processes occurring at the vertex. It addresses the challenge of identifying and measuring independent particles within this complex environment, where multiple tracks overlap. The implications of this advancement are profound, as it provides a means to delve deeper into the characteristics and properties of these particles, ultimately enabling a better understanding of the incoming neutrino kinematics. In the realm of neutrino physics, where accurate measurement of kinetic energy, direction, and particle counts is essential, the deep learning-based image decomposition approach offers a transformative opportunity. The high resolution achieved in particle kinetic energy is particularly noteworthy. This level of precision has the potential to significantly enhance our ability to explore oscillations~\cite{Pontecorvo-1957-cp,Fukuda-1998-evidence,Ahmad-2002-direct}.
By providing a clearer and more detailed view of the interaction vertex, this work lays the foundation for improved accuracy in neutrino-related measurements. Finally, the results presented in Section~\ref{sec:comparison}, evaluated on a statistically independent dataset of neutrino interactions under a selected nuclear model, underscore excellent generalisation performance of the method for the selected case study. Besides, it proves the impact of our deep-learning approach on future high-precision long-baseline experiments, with the capacity to reduce associated systematic errors, avoid model dependence, and improve the neutrino energy resolution, which directly influences the sensitivity towards potentially discovering leptonic charge-parity violation and measuring neutrino oscillation parameters~\cite{ershova2023role}.

Integrating a generative model of individual particles that can be fine-tuned on detector data (e.g., charged-particle beam tests) represents an exciting avenue for further enhancement of image decomposition. It is worth noting that the intrinsic differentiability of the generator exhibits a remarkable and atypical characteristic within the realm of data analyses in particle physics, which is currently a thriving area of research~\cite{gasiorowski2023differentiable,grinis2022differentiable, Alonso-Monsalve2020b, roussel2022applications}. This property facilitates the construction of a statistical likelihood, permitting the calculation of systematics and precise quantitative measurements. Thus, this approach not only enables improved accuracy in the reconstruction process but also allows for the computation and propagation of errors to downstream analysis methods. It is an example of how interdisciplinary approaches that combine computer vision and particle physics can mutually benefit from each other. As a result, the implications of this work extend beyond the confines of both fields, offering a synergistic approach between particle physics and computer vision.

In conclusion, our paper exhibits the potential of deep learning and computer vision techniques in advancing our understanding of particle physics experiments. The ability to address overlapping-sparse images in the context of neutrino physics holds the promise of not only enhancing our knowledge of fundamental particles but also pushing the boundaries of computer vision research, offering a powerful tool for image decomposition in various applications. As future work, we aim to analyse different network architectures more comprehensively and evaluate the system's resilience to adversarial attacks for improved robustness.

\section{Methods}
\label{sec:methods}

\subsection{Simulated datasets}
\label{sec:simulation}

Simulated datasets are generated within a cubic uniform plastic scintillator detector comprising a grid of $9 \times 9 \times 9$ identical voxels, each possessing a volume of 1 cm$^{3}$. The coordinates x, y, and z, as well as the angles phi and theta, are discussed with respect to the standard spherical coordinate system, where the centre of the volume is located at (0, 0, 0), and phi is defined as the angle orthogonal to the z-axis. 
The simulated detector bears a resemblance to 3D imaging detectors such as fine-grained plastic scintillators or liquid argon time projection chambers (LArTPCs), common in long-baseline neutrino experiments~\cite{Blondel_2020, Alekseev_2023, app11062455}. 
Particles are uniformly generated within the central voxel, with their initial directions isotropic and initial kinetic energies uniformly distributed. The simulation process consists of three consecutive steps:
\begin{enumerate}
    \item \textbf{Energy loss simulation}: employing the Geant4 toolkit~\cite{geant4_1, geant4_2, geant4_3} to model particle propagation within the detector and compute the local energy loss along the particle trajectory. The energy loss calculation accounts for Birks' quenching effect, with a fixed correction coefficient of 0.126 mm/MeV applied uniformly to all charged particles.
    
    \item \textbf{Detector response simulation}: transforming data obtained in the preceding step into ``signal'', specifically scintillation light, that the instrument can detect. We implement a conversion factor to directly scale the energy loss from its physical unit (MeV) to a signal unit (photon electron, p.e.). The chosen scaling factor is set at 100 p.e./MeV. Additionally, we incorporate a detector effect in our simulation, known as ``crosstalk'', which accounts for light leakage into neighbouring voxels. The leakage fraction per face is established at 3\%.

    \item \textbf{Event Summary}: gathering the simulated photon electron signals within each voxel and compiling event information into a format compatible with neural network processing.
\end{enumerate}

\begin{figure*}[htb] 
\centering
\includegraphics[width=0.9\textwidth]{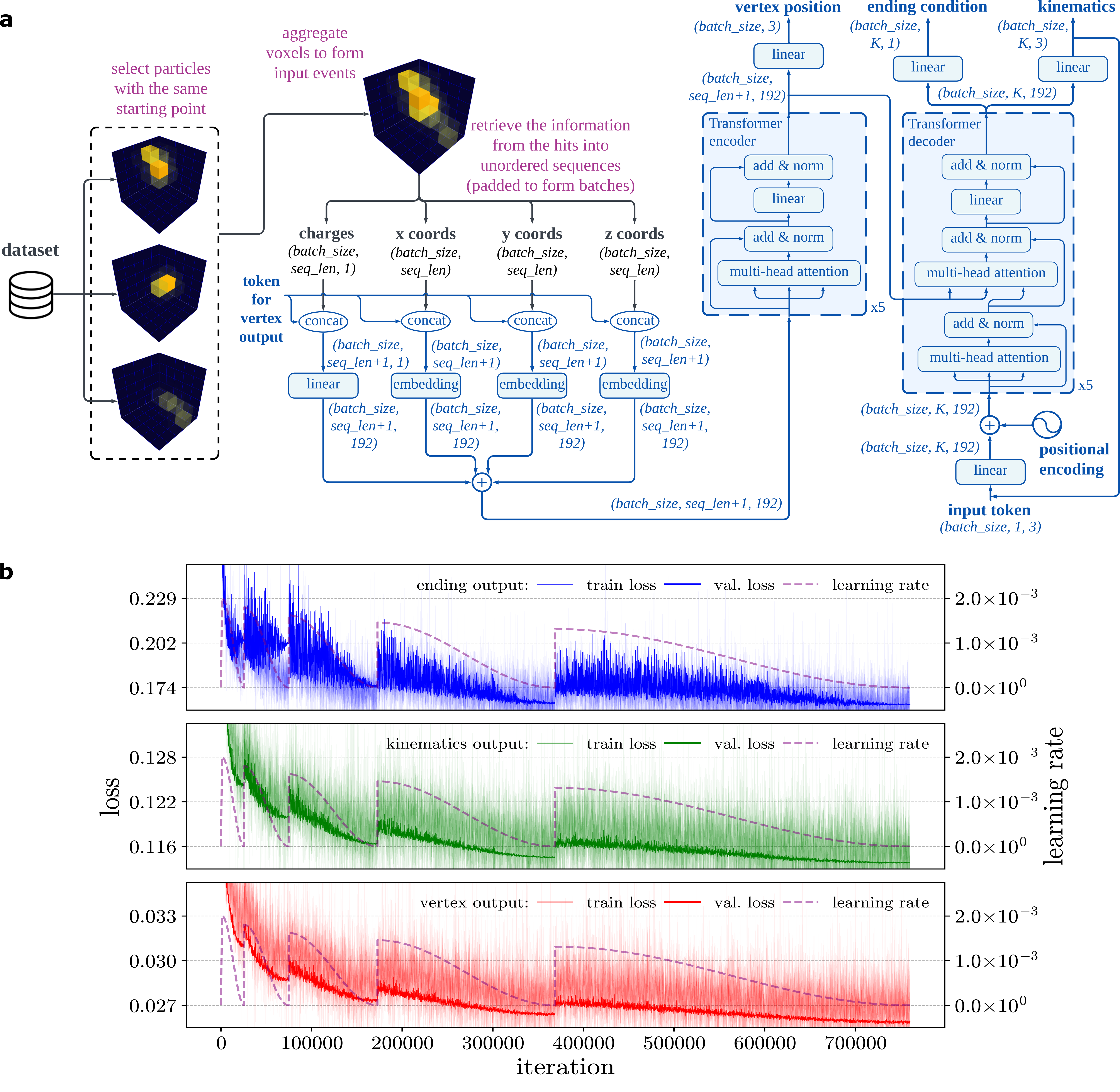}
  \caption{\label{fig:fitting}
  \textbf{Architecture and training curves of the decomposing transformer.}
  \textbf{a} Left to right: at training time, a variable number of particles with similar (within $\sim$0.2 mm) initial positions are selected from the dataset and combined to create an event, and multiple events form a training batch. The input voxel data (comprising energy loss and spatial coordinates) is subsequently processed and passed through the decomposing transformer. This transformer initially generates a prediction for vertex positions and subsequently provides estimates for kinematic parameters and termination conditions for each particle in the events. The diagram illustrates the evolving tensor shapes at various stages of the processing.
  \textbf{b} Training and validation curves for the different outputs of the network, showing a smooth convergence of the model (this plot corresponds to the model from Section~\ref{sec:initial_case}. Similar curves are observed for the model from Section~\ref{sec:nuclear_clusters}). The learning rate schedule used is appreciated by looking at the dashed purple lines. The parentheses indicate the tensor dimensions at each stage.}
\end{figure*}

The simulated particles include four distinct types: muons ($\mu^{-}$), protons, ionised deuterium nuclei ($D^{+}$), and tritium nuclei ($T^{+}$). A total of five million events represent each type. The initial kinetic energies for these particles are determined based on the CC$0\pi$ vertex-activity signals we anticipate observing: muons exhibit kinetic energies ranging from 300 to 1000 MeV; protons exhibit kinetic energies between 5 and 60 MeV, with the majority coming to a halt within one or two voxels, thereby contributing to an energy concentration in the vicinity of the vertex; ionised deuterium and tritium nuclei possess kinetic energies between 10 and 60 MeV, resulting in a more pronounced energy concentration in the vertex region due to their higher energy loss (dE/dx).

To replicate CC$0\pi$ neutrino interaction events, the simulated particles are organised into two distinct event types during the training process, contingent upon the specific scenario under investigation:
\begin{itemize}
    \item \textbf{Scenario 1} (Section~\ref{sec:initial_case}): Each event consists of a single muon, accompanied by an arbitrary number of protons, with the number of protons being randomly selected from 1 to 5.
    \item \textbf{Scenario 2} (Section~\ref{sec:nuclear_clusters}): This dataset includes one muon, an arbitrary count of protons (ranging from 0 to 4), and an arbitrary number of ionised deuterium (ranging from 0 to 1) and tritium (ranging from 0 to 1) nuclei.
\end{itemize}

To group particles by starting position, the central voxel was virtually divided into a grid of 125,000 smaller sub-voxels, each with dimensions of 0.2$\times$0.2$\times$0.2 mm$^3$. We consider two particles to start from the same position if their starting point falls within the same virtual sub-voxel. Consequently, each dataset contains approximately 40 particles originating from each sub-voxel. To form events, we randomly select particles starting from the same position, sum their voxel charges, and introduce a random shift of $\pm$1 voxel in every direction to the entire event. This random shifting enables the algorithm to learn that the interaction vertex position can be located anywhere within the central 3$\times$3$\times$3 voxels of the original 9$\times$9$\times$9 input volume (assuming a pessimistic prior vertex reconstruction for modern detectors to ensure robustness to the algorithm). The outer surface of the final volume is omitted to prevent incomplete events due to the shifting, resulting in a final event shape of 7$\times$7$\times$7 voxels. We choose a single simulated event per sub-voxel for the validation and test sets, resulting in a fixed number of 125K validation and 125K testing events. Just an example, for the proton case, the remaining approximately 37 protons per sub-voxel are allocated for training, yielding $\sim$510K combinations per sub-voxel (combinations of 1 to 5 protons without repetition or order from a group of 35) and a total of $\sim$63 billion (for all the 125K sub-voxels) possible combinations of training events.

The dataset utilised in Section~\ref{sec:comparison} was generated using the NEUT neutrino generator version 5.7.0~\cite{Hayato_2021}, with the following neutrino parameters: muon neutrinos only, initial energy uniformly distributed between 0 and 1.0 GeV, initial position uniformly distributed inside the central cube, and initial direction pointing to the Z-axis (in our detector reference system). The simulated interaction topology is CC$0\pi$ (i.e., with one muon and no pions). To ensure a fair comparison to the events used for training, only events with proton energies ranging from 5 to 60 MeV and no neutrons or photons are included in this study.

\subsection{Transformer for sparse-image decomposition}
\label{sec:transformer}

The objective of the decomposing algorithm is to take an image as input and output global features (i.e., the 3D position of the interaction vertex in our case study), along with specific properties of each constituent component in the image. In our neutrino-interaction case, those properties are (per particle): initial kinetic energy, initial direction specified in spherical coordinates ($\theta$ and $\phi$), and particle type (optional if all particles share the same type).

The neural network architecture employed for the decomposing algorithm is a transformer model~\cite{vaswani2017attention} (illustrated in Fig.~\ref{fig:fitting}a). The transformer implemented has a hidden size of 192 and consists of 5 encoder layers and 5 decoder layers. Additionally, it utilises 16 attention heads. We selected Lamb~\cite{you2020large} as the optimiser. The total number of trainable parameters of the model is 10,393,165. The effective batch size used was 2048, achieved by accumulating gradients over an actual batch size of 512 with an accumulation factor of 4. Other optimiser parameters were set as follows: $\beta_1=0.9$, $\beta_2=0.999$, and weight decay of 0.01. The hyperparameters were chosen via a grid search process. In order to facilitate proper convergence of the model, a learning rate warm-up strategy was employed. During the first 20 epochs, the learning rate gradually increased until it reached an upper limit of 0.002. Subsequently, the learning rate was reduced using a cosine annealing with a warm restart schedule. The lower limit of the learning rate was set to 1000 times smaller than the upper limit. The number of epochs before the first restart was set to 400, and this number was multiplied by a factor of 2 after each restart. Besides, at each warm restart, the upper limit of the learning rate was multiplied by a decay factor of 0.9. The training and validation curves illustrating the model performance can be observed in Fig.~\ref{fig:fitting}b. The same architecture was used to obtain the results presented in Sections~\ref{sec:initial_case}~and~\ref{sec:nuclear_clusters}. The training was carried out using an NVIDIA V100 GPU, with Python 3.10.12, PyTorch 2.0.0~\cite{NEURIPS2019_9015}, and PyTorch Lightning~\cite{Falcon_PyTorch_Lightning_2019} used for the implementation.

\subsection{Generative adversarial network for fast elementary-particle simulations}
\label{sec:gan}

\begin{figure*}[htb] 
\centering
\includegraphics[width=0.9\textwidth]{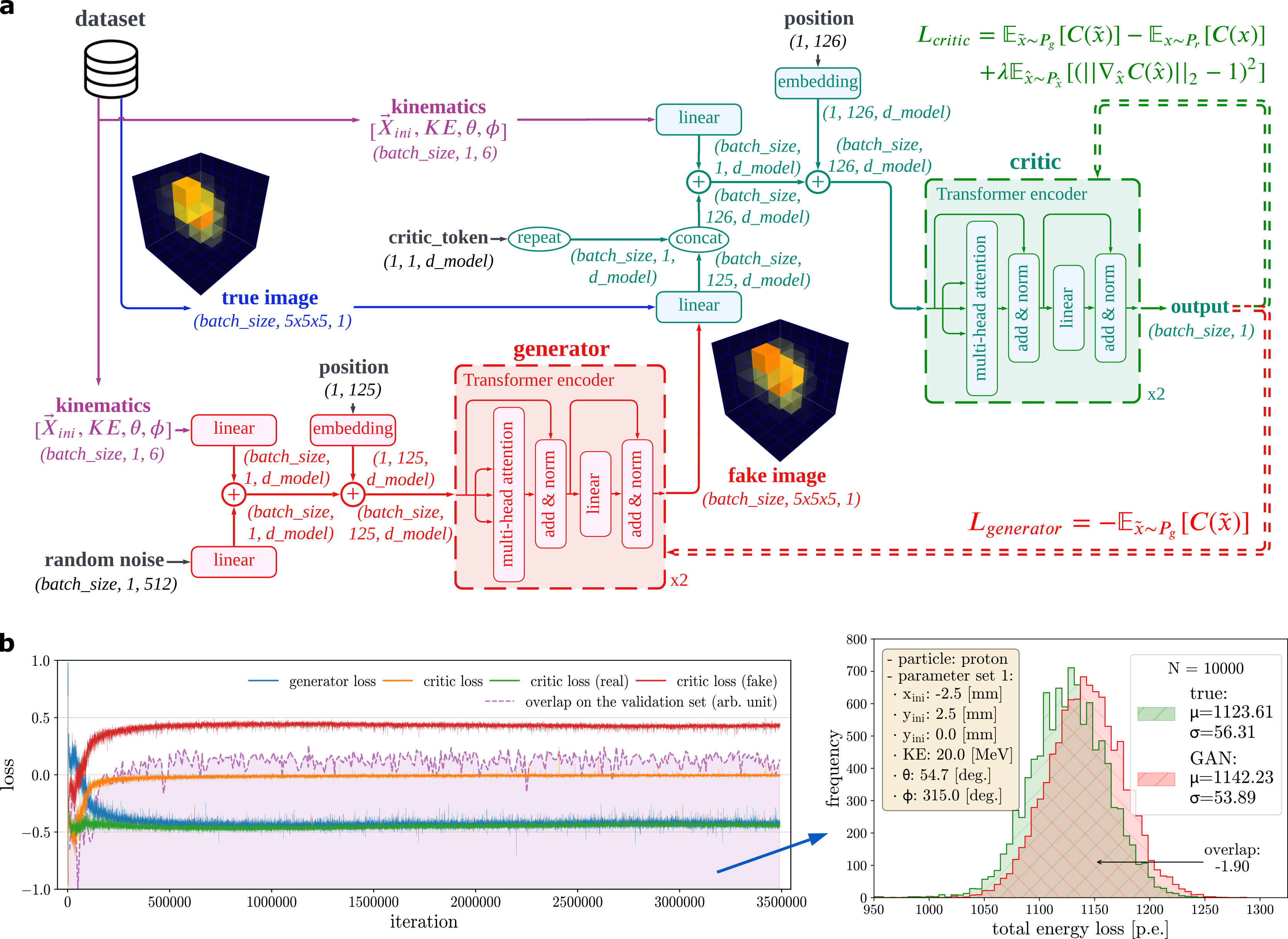}
  \caption{\label{fig:gan}
  \textbf{Architecture and training curves of the generative adversarial network.}
  \textbf{a} At training time, candidate kinematics are combined with random noise vectors and passed to the generator (highlighted in red), which transforms them into a synthetic image. The task of the discriminator (depicted in green) is to distinguish between true and synthetic images. The diagram illustrates the changes in tensor shapes during various stages of this process, as well as the loss functions minimised during the training.
  \textbf{b} Training and validation curves display the performance of the different components within the GAN used for proton generation (similar curves are observed for other particles). The purple area denotes the average overlap between the GAN-generated and simulated distributions for five randomly selected parameter sets, each consisting of 10,000 real and synthetic events. On the right-hand side, we illustrate the true and GAN-generated distributions, highlighting their overlap for a randomly selected parameter set obtained from an arbitrary model checkpoint. It serves as an indicator of the overall performance of the GAN, with larger values indicating superior performance.}
\end{figure*}

To assess the performance of the vertex-activity fitting, one could generate candidate images on the fly using the transformer predictions as input (which can be extremely time-consuming due to the complexity of the simulation software) or use a pre-generated library (e.g., the training dataset) and, for each particle predicted by the fitting method, find the closest match in terms of kinematics from the library. However, this latter process might also be very time-consuming, especially when dealing with a large library. The purpose of needing a sizeable library is to ensure a correct validation of the method by sampling the entire distribution. In this article, an alternative approach is adopted, leveraging a generative model to replicate the simulation process. Once trained, the generative model efficiently produces particle images based on the provided kinematics as input, eliminating the need for time-consuming library searches. This approach enables rapid generation of particle images, bypassing the conventional procedure of matching particles with a pre-generated library.

The chosen generative model in this study is a generative adversarial network (GAN)~\cite{10.5555/2969033.2969125}, specifically a conditional GAN (cGAN)~\cite{mirza2014conditional} integrated into the Wasserstein GAN with Gradient Penalty (WGAN-GP) framework~\cite{arjovsky2017wasserstein, gulrajani2017improved}. This model is conditioned on input kinematic parameters and a Gaussian noise vector to generate particle images, allowing for controlled and realistic image synthesis. WGAN-GP addresses some of the limitations of standard GANs, enhancing their stability and convergence. It employs the Wasserstein distance (also known as Earth Mover's distance) as the loss function and enforces a constraint such that the gradients of the critic's output with respect to the inputs have a unit norm (gradient penalty). The implementation of WGAN-GP utilises a transformer encoder as the architecture for both the generator and the critic (revealed in Fig.~\ref{fig:gan}a), each with 2 encoder layers, 8 attention heads, and a hidden size of 64. The generator takes the kinematics as input to generate particle images, while the critic assesses the quality of the generated images. Both the generator and the critic have the same number of trainable parameters, which amounts to 141,569. The RMSprop optimiser~\cite{tieleman2012lecture} is employed for both the generator and the critic, with a batch size of 32 and a learning rate of $5\times10^{-4}$ in both cases. The critic is updated 5 times for every update of the generator. Figure~\ref{fig:gan}b presents the training curves and the validation metric. Again, the training process utilised an NVIDIA V100 GPU, with Python 3.10.12, PyTorch 2.0.0~\cite{NEURIPS2019_9015}, and PyTorch Lightning~\cite{Falcon_PyTorch_Lightning_2019} used for the implementation.

\subsection{Gradient-descent minimisation of the image parameters}
\label{sec:gradient_descent}

The initial variant involves the execution of $n$ independent gradient-descent minimisation processes. Each minimisation employs a unique, predetermined noise seed for the GAN. The primary objective of this approach is to identify the optimised parameters that yield a reconstructed image with the closest resemblance to the target image. This operational procedure is elucidated in Algorithm~\ref{alg:gradient_descent1}. This optimisation endeavour aims to minimise the mean-squared error, serving as the pertinent loss function.

\begin{algorithm}[htb]
  \caption{Parameter optimisation via gradient descent for best-fit image}
  \label{alg:gradient_descent1}
  \SetKwInput{Input}{Input}
  \SetKwInput{Output}{Output}
  \Input{\\
    - $I_{tgt}$: target image\\
    - $\theta_{tra}$: parameters predicted by transformer\\
    - $n$: $\#$ of independent gradient-descent runs\\
    - $m$: $\#$ of iterations for each gradient-descent run\\
    - $z$: input noise for each GAN run (length of $n$)\\
    - $lr$: learning rate
  }
  \Output{\\
    - $\theta^*$: best-fit parameters
  }
  $\ell_{\text{min}} \leftarrow \infty$;\Comment{minimum loss value}\\
  $\theta^* \leftarrow \varnothing$\;
  \For{$i \leftarrow 1$ to $n$}{
    $\theta \leftarrow \theta_{tra}$;\Comment{initial params}\\
    $\mathcal{O} \leftarrow \text{Adam}(\theta, \text{lr})$;\Comment{optimiser initialisation}\\
    \For{$j \leftarrow 1$ to $m$}{
      $\mathcal{O}.zero\_grad()$\;
      $I_{reco} \leftarrow \text{GAN}(\theta, z[i])$;\Comment{GAN on all particles}\\
      $\ell \leftarrow \text{lossf}(I_{reco}, I_{tgt})$;\Comment{loss calculation}\\
      $\ell.backward()$\; 
      $\mathcal{O}.step()$;\Comment{update params}\\
      \If{$\ell < \ell_{\text{min}}$}{
        $\ell_{\text{min}} \leftarrow \ell$\;
        $\theta^* \leftarrow \theta$\;
      }
    }
  }
  \Return{$\theta^*$}\;
\end{algorithm}

\begin{algorithm}[htb]
  \caption{Parameter optimisation via gradient descent for likelihood inference}
  \label{alg:gradient_descent2}
  \SetKwInput{Input}{Input}
  \SetKwInput{Output}{Output}
  \Input{\\
    - $I_{tgt}$: target image\\
    - $\theta_{tra}$: parameters predicted by transformer\\
    - $n$: $\#$ of GAN runs per iteration\\
    - $m$: $\#$ gradient-descent iterations\\
    - $lr$: learning rate
  }
  \Output{\\
    - $\theta^*$: best-fit parameters
  }
  $\ell_{\text{min}} \leftarrow \infty$;\Comment{minimum likelihood value}\\
  $\theta^* \leftarrow \varnothing$\;
  $\theta \leftarrow \theta_{tra}$;\Comment{initial params}\\
  $\mathcal{O} \leftarrow \text{Adam}(\theta, \text{lr})$;\Comment{optimiser initialisation}\\
  \For{$j \leftarrow 1$ to $m$}{
    $\mathcal{O}.zero\_grad()$\;
    $\tau \leftarrow \varnothing$;\Comment{template to be filled}\\
    \For{$i \leftarrow 1$ to $n$}{
      \parbox[t]{6.75cm}{\raggedleft $\tau \leftarrow \tau + \text{GAN}(\theta)/n$;\Comment{GAN on all particles;\\noise is random}}\\
    }
    $\ell = -2ln\mathcal{L}(I_{tgt}, \tau)$;\Comment{likelihood calculation}\\
    $\ell.backward()$\; 
    $\mathcal{O}.step()$;\Comment{update params}\\
    \If{$\ell < \ell_{\text{min}}$}{
      $\ell_{\text{min}} \leftarrow \ell$\;
      $\theta^* \leftarrow \theta$\;
    }
  }
  \Return{$\theta^*$}\;
\end{algorithm}

In the context of the second variant, the primary objective differs from that of the first. Here, the aim is not to identify the specific parameters and noise seed that generate an image most closely resembling the target image. Instead, the focus is on mitigating the inherent stochasticity of the method, enabling the calculation of uncertainties and the utilisation of the results in subsequent physics investigations. This second variant (Algorithm~\ref{alg:gradient_descent2}) bears resemblances to the previous one, albeit with certain distinctions. Firstly, the optimisation process involves a single gradient-descent optimisation applied to the parameters. However, in each iteration, $n$ independent reconstructed events are generated using the GAN, and these events are subsequently averaged to create a unified template image. This template image serves as the basis for minimising a likelihood function $\mathcal{L}$. The likelihood $\mathcal{L}$ is computed as a product of the Poisson likelihood ratios for the energy loss in each voxel, comparing the template (expected) and target (observed) images. This likelihood formulation is instrumental in quantifying the goodness of fit between the two images (inspired by~\cite{BAKER1984437, PhysRevD.103.112008}):

\begin{equation}
-2\ln \mathcal{L} = 2\sum_{i}^{N}\left(\text{obs}_i \cdot \ln\left(\frac{\text{obs}_i}{\text{exp}_i}\right) + \left(\text{exp}_i - \text{obs}_i\right)\right)
\end{equation}

where $obs$ is the observation (target image, presented as a flattened 1D vector), $exp$ the expectation (template, also as a flattened 1D vector), and $N$ denotes the total number of voxels in the image.

To provide further insight, the optimisation process of both variants employs the Adam~\cite{Kingma-2014-adam} optimiser with specific learning rates tailored to the various parameter categories: a learning rate of 0.005 is employed for the vertex-position parameters, 0.05 for the kinetic energy, and 0.2 for the direction. These values were chosen based on the discerned patterns in the transformer results, as illustrated in Fig.~\ref{fig:results}. Furthermore, the optimisation process encompasses 200 independent gradient-descent minimisations for the first variant, or the number of GAN runs per iteration for the second variant (designated as $n$). Each of these minimisation processes is iterated for 50 times the number of reconstructed particles present in the event under consideration.

\section{Data availability}

The datasets used to train and test our models are publicly available at \url{https://doi.org/10.5281/zenodo.10075666}.

\section{Code availability}

All code used to implement the methods and reproduce the findings presented in this paper is publicly available at \url{https://github.com/saulam/NeutrinoVertex-DL}.

\section*{Acknowledgements}

Part of this work was supported by the SNF grant PCEFP2\_203261, Switzerland.

\bibliographystyle{ieeetr}
\bibliography{biblio}

\end{document}